\documentclass{article}
\usepackage{ijcai20}

\usepackage{times}
\usepackage{soul}
\usepackage{url}
\usepackage[hidelinks]{hyperref}
\usepackage[utf8]{inputenc}
\usepackage[small]{caption}
\usepackage{graphicx}
\usepackage{amsmath}
\usepackage{amsthm}
\usepackage{booktabs}
\usepackage{algorithm}
\usepackage{algorithmic}
\usepackage{balance}
\usepackage{xcolor}

\title{Is Attention All What You Need ? - An Empirical Investigation on  Convolution-Based Active Memory and Self-Attention}

\author{
Thomas Dowdell\and
Hongyu Zhang
\affiliations
The University of Newcastle, NSW, Australia
\emails
tomjamesdowdell@gmail.com,
hongyu.zhang@newcastle.edu.au
}

\begin{document}

\maketitle

\begin{abstract}
  The key to a Transformer model is the self-attention mechanism, which allows the model to analyze an entire sequence in a computationally efficient manner. Recent work has suggested the possibility that general attention mechanisms used by RNNs could be replaced by active-memory mechanisms. In this work, we evaluate whether various active-memory mechanisms could replace self-attention in a Transformer. Our experiments suggest that active-memory alone achieves comparable results to the self-attention mechanism for language modelling, but optimal results are mostly achieved by using both active-memory and self-attention mechanisms together. We also note that, for some specific algorithmic tasks, active-memory mechanisms alone outperform both the self attention and a combination of the two. 
\end{abstract}

\section{Introduction}

The previous state-of-the-art sequence model, the recurrent neural network, has been largely supplanted by the Transformer model \cite{vaswani2017}, which is primarily built atop a self-attention mechanism. Given a task to train upon, the self-attention mechanism focuses on one token per attention head within the entire sequence at each time-step; the key to the self-attention mechanism’s success is the mechanism’s ability to learn which token within the entire sequence to focus on in order to achieve the best results.

The self-attention mechanism has proven successful on a variety of natural language processing tasks, but has not achieved ubiquitous success. The authors of \cite{kaiser2016} pointed out that an attention mechanism would likely struggle to solve a task which required a model to focus on multiple tokens at a given time-step. Further, the authors of \cite{kaiser2015} recommended that an attention mechanism could be replaced by active-memory to alleviate these concerns.

\begin{figure}[!h]
\includegraphics[clip,trim={6.5cm 7.7cm 11cm 8.7cm},width=8.6cm]{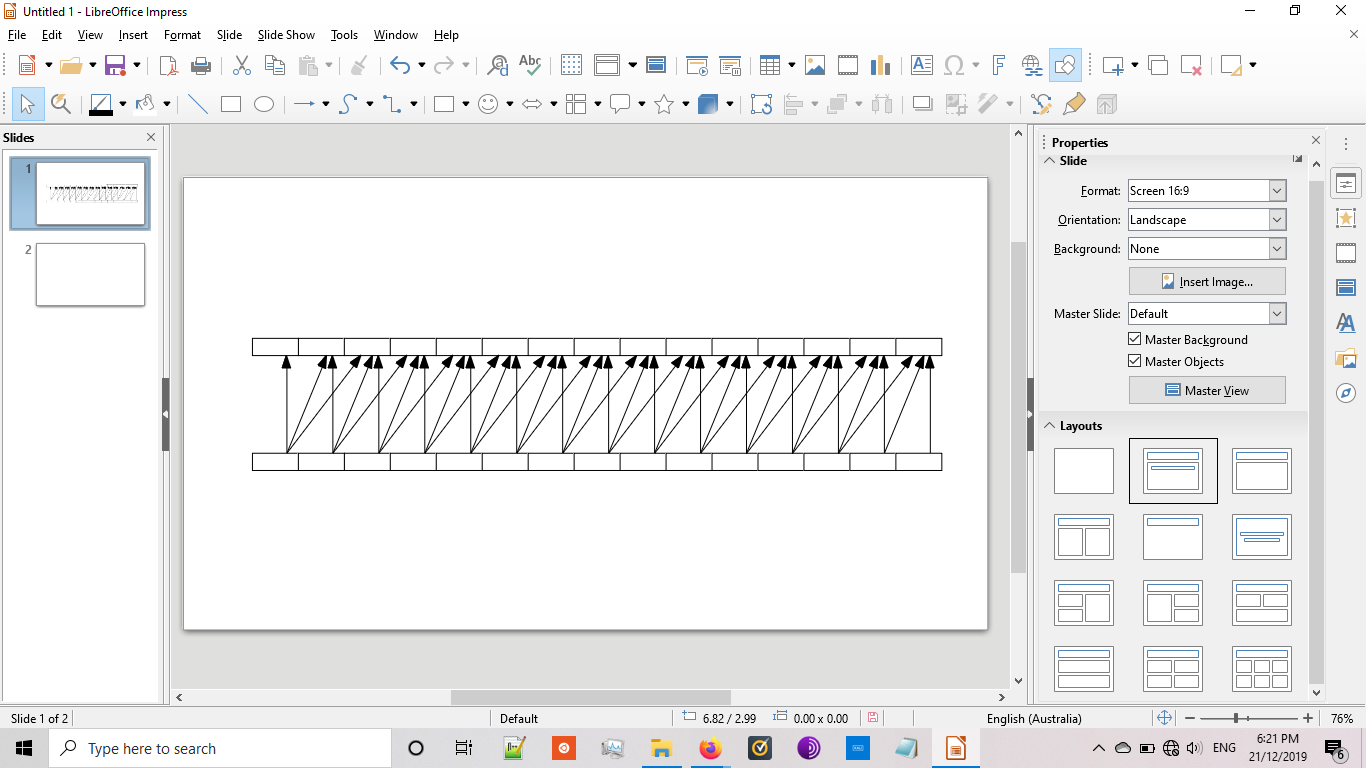}
\vspace{-6pt}
\caption{The active memory mechanism. In this case, the active-memory is implemented in a unidirectional manner, with a kernel size 3.}
\label{fig:arch}
\vspace{-6pt}
\end{figure}

Unlike attention, active-memory allows a model to access and change any and all elements of its memory at each time-step. 
The active-memory mechanism can access more than one element at each time step. 
In \cite{kaiser2016}, the authors used an active-memory system to translate English to French, and was capable of outperforming an RNN model, both with and without an attention mechanism.

Motivated by the success of attention mechanism \cite{vaswani2017} and active-memory \cite{kaiser2016}, in this paper we investigate the Transformer's self-attention mechanism in comparison to a variety of active-memory mechanisms. We experiment on two types of tasks: the language modeling task and a set of algorithmic tasks.

For the language modelling task, the self-attention mechanism out-performs an active-memory mechanism used alone by a slim margin. However, a combination of both self-attention and active-memory reliably outperform both mechanisms used alone.

We also evaluated the self-attention mechanism and various active-memory mechanisms on a variety of algorithmic tasks, which can also be expressed as a sequence modeling task. Across most of the algorithmic tasks tested, the active-memory mechanisms achieve equal, or superior, results to a traditional self-attention mechanism. This would appear to vindicate the hypothesis stated by \cite{kaiser2016}, suggesting that the nature of the attention mechanism does indeed limit the effectiveness and accuracy of the model.
Finally, we note that, for several algorithmic tasks, the mere addition of the self-attention mechanism hinders results; the active-memory mechanism alone outperforms a combination of the two separate mechanisms. This raises an unsolved problem; it would appear that, for deep learning sequence models, there is still no unambiguous model that can optimally solve all possible problems.

\section{Related Work}

The Transformer model \cite{vaswani2017} is built with two separate modules, the self-attention mechanism and the feedforward mechanism, which are stacked atop each other for multiple layers. The feedforward mechanism is an intra-sequence analysis, where the output for each token in the sequence is dependent only on the token at the same time-step, and independent of all other time-steps. On the other hand, the self-attention mechanism is an inter-sequence analysis, where the output for each time-step is dependent upon the entire sequence. The self-attention mechanism is defined, mathematically, as:

\[Q_{t}, K_{t}, V_{t} = x_{t} \]
\[y_{t} = concat(head_{1,t}, head_{2,t}, …, head_{n,t})W_{o} \]
\[head_{i} = Attention(Q_{t}W_{i}^Q, K_{t}W_{i}^K, V_{t}W_{i}^V) \]
\[Attention(Q, K, V) = softmax(\frac{Q_{t}K_{t}^T}{\sqrt d_k}) V_{t} \]
\[W_{o} \epsilon R^{d_k * k, d}, W_{i}^{K,Q,V} \epsilon R^{d, d_k * k} \]

The feed-forward module is defined as:

\[y_{t} = W_{l,2}(max(W_{l,1}x_{t} + b_{l,1}, 0)) + b_{l,2} \]
\[W_{l,2} \epsilon R^{d, d * 4}, W_{l,1} \epsilon R^{d * 4, d} \]

The Transformer model, and its variants \cite{transformerxl}, have achieved remarkable results across a variety of natural language processing tasks since its inception \cite{albert} \cite{bert} \cite{xlnet} , and are currently investigated heavily by both academia and industry.

The Neural GPU \cite{freivalds} \cite{kaiser2016} \cite{kaiser2015}, which introduced an active-memory model, achieved impressive algorithmic results in \cite{kaiser2015}, and also achieved impressive machine translation results in \cite{kaiser2016}. A Neural GPU contains a CGRU (Convolution Gated Recurrent Unit) module which is iterated repeatedly. This allows the entire sequence to be analyzed in a parallelizable and computationally efficient manner. The CGRU module is defined as:

\[u = sigmoid(U_1 * x + B_1) \]
\[r = sigmoid(U_2 * x + B_2) \]
\[y = u \otimes x + (1 - u) \otimes tanh(U_0 * (r \otimes x) + B_0)) \]
 
\noindent
where \textit{U * x} refers to applying a convolutional operator over \textit{x}, using \textit{U} as a trainable kernel bank and \textit{B} is a trainable bias vector. The CGRU has, since its introduction, been used in other models \cite{one_shot_draw}.

Convolutional operators are traditionally used for image processing \cite{historyofdeeplearning}, and have also been used in relation to sequential analysis in previous papers \cite{yang2019} \cite{wu2019} \cite{gehring2017} \cite{glu2017}. To the best of our knowledge they have not been used explicitly to replace, or augment, the self-attention mechanism. The first sequence-to-sequence model, based on convolutional operators, was, to the best of our knowledge, introduced in \cite{gehring2017}, which replaced the then-traditional LSTM block with a series of convolutions and gated convolutional networks [13], and outperformed RNN-based models in terms of both speed and accuracy. However, the model introduced in \cite{gehring2017} was followed shortly afterwards by the Transformer model, which outperformed the convolutional-based model.

The convolutional self-attention network \cite{yang2019} was recently introduced, and bares a passing similarity to the traditional convolutional operator described in this paper. The layer of the convolutional self-attention is similar to a traditional self-attention mechanism, but where the key and value tensors are calculated as:

\[K^h = (K^h_{i - M/2}, ..., K^h_{i}, ..., K^h_{i + M/2}) \]
\[V^h = (V^h_{i - M/2}, ..., V^h_{i}, ..., V^h_{i + M/2}) \]

From this point, the convolutional self-attention mechanism acts in an identical manner to the traditional self-attention mechanism. This is in direct comparison to the convolutional operator described in this paper, which explicitly avoids the use of the self-attention mechanism and relies entirely on a purely convolutional operator.

\section{Approach}

In this paper, we investigate whether various active-memory mechanisms could replace self-attention in a Transformer. We also evaluate the combination of self-attention and active-memory mechanisms for language modelling tasks. All the active-memory mechanisms introduced in this paper were inspired by the Neural GPU, as introduced in \cite{kaiser2015}. The key allure of the Neural GPU is that the inputs of each time-step can be analyzed and altered, and we were inspired to apply a similar form of sequence modelling alongside a self-attention mechanism.
We describe various convolution-based active-memory mechanisms in this section.
 
\subsection{The Convolutional Operators}

\subsubsection{The Traditional Convolutional Operator } 

The first, and most simple, active-memory mechanism is the simple convolutional operator. The traditional convolutional operator was formally defined in \cite{bai}. If the task requires the sequence to be analyzed in a unidirectional manner, such as the case for language modelling, then a zeros-vector of size \textit{k – 1} is concatenated to the left of the input tensor so that, for the \textit{nth} output token, the model only has access to the first \textit{n} input tokens. This feature is crucial to avoid allowing the model ‘seeing’ forward through the sequence and having access to information that the model, in practice, would not yet have. This has an identical function to the masking operation of the self-attention mechanism.

If the task can be analyzed in a bidirectional manner, then the model uses a convolutional filter using the SAME-padding, which allows for the vector to maintain its shape throughout the convolutional operator. However, when the convolutional operator is performed in this manner, the token at time-step \textit{t} is dependent on the input tokens \textit{h\textsubscript{[t-k/2,t+k/2]}}, where \textit{k} is the kernel size.

The primary flaw of a convolutional operator, in comparison to a self-attention mechanism, is that, given \textit{n} layers where each kernel has a \textit{k} kernel size, each token can only see \textit{k * n – n + 1} or \textit{k/2 * n - n + 1} time-steps across for unidirectional and bidirectional tasks respectively. For example, in our experiments on language modeling (Section 4), the kernel size was set to 20 and was iterated over 8 layers. Therefore, at each time-step \textit{t}, the final output is capable of analyzing the input from 153 previous time-steps, well above the average sequence-size (90 tokens) in the dataset.
The self-attention mechanism, in comparison, can see across a theoretically infinite context size, even using only a single layer. Given this information, the self-attention mechanism is capable of handling theoretically greater long-term dependencies than the active-memory mechanism. However, in practice, the ability of an active-memory mechanism to access and change its entire memory could overcome this limitation.

The convolutional operator is assisted further by the fact that the convolutional operator’s complexity grows linearly with the sequence size, while the self-attention mechanism’s complexity grows quadratically.

Numerous papers have noted that, while Transformers are parallelizable and capable of capturing long-range dependencies, the Transformer network suffers from the inability of model tokens in a recurrent manner \cite{wang2019} \cite{hao}. This is in direct comparison to traditional RNN models, which can capture long-range dependencies, but can struggle to capture long-range dependencies. The use of active-memory, in theory, would accomplish this task, given that the output at time-step \textit{t} \textit{h\textsubscript{t}} is dependent of the inputs \textit{x\textsubscript{[t-k,t]}} where k is the kernel size. Therefore, this operation can, in theory, model recurrence. We did not explicitly test whether this does model recurrence in practice, but will focus on this in future work.

The convolutional operator is followed by the ReLU activation function.

\subsubsection{The Persistent-Convolutional Operator}

The Persistent-Convolutional operator is similar to the traditional convolutional operator described above, except that the zeros vector is replaced by the a trainable vector of identical shape to the zeros vector. This allows the operator to, identical to the traditional convolutional operator, maintain an identical shape across the convolution. To keep parameterization to a minimum, the same persistent vector is used across all convolution operators in the entire model. The persistent-convolutional operator is defined as:

\[p \in W^{kernel\_size - 1, hidden_size} \]
\[x = [p, x],   y = W * x + b \]

\noindent
where \textit{[.,.]} denotes the concatenation function and \textit{p} is the trainable persistent vector. Persistent vectors have been used previously in language modelling tasks \cite{sukbaatar2019}, but never as an augmentation for convolutional operators, as far as we know.

If the model is to be analyzed in a bidirectional manner, rather then a unidirectional manner, then the persistent-convolutional operator can be redefined as:

\[p_1, p_2 \in W^{(kernel\_size - 1)//2, hidden_size} \]
\[x = [p_1, x, p_2] \]

The use of a persistent vector allows for the model to have a permanent memory that, given the fact the vector is trainable, can be expressed in an optimal manner for the model. This is the equivalent of a permanent memory for the deep learning model.

\subsubsection{The Highway-Convolutional Operator}

The Highway-Convolutional operator is based on the highway network architecture [5], which can be defined as:

\[a = U_0 * x + B_0 \]
\[b = sigmoid(U_1 * x + B_1) \]
\[y = a \otimes b + x \otimes (1 - b) \]

The key allure of the highway network, as described in \cite{srivastava2015}, is the fact that a highway network can be trained for a large number of layers, even hundreds of layers, because information can pass, unimpeded, across each layer. The authors of \cite{srivastava2015} described these paths as 'information highways'. The use of these 'information highways' allows information to pass through the self-attention mechanism in an equally efficient manner.

In this paper, we use the hard-sigmoid function \cite{kaiser2016} to stabilize gradients, which is defined as: 

\[ y = max(0, min(1, 1.2 * sigmoid(x) - 0.1)) \]

\begin{figure}
\begin{center}
\vspace{-12pt}
\includegraphics[clip,trim={10.5cm 3.72cm 14cm 4.8cm},width=0.4\textwidth]{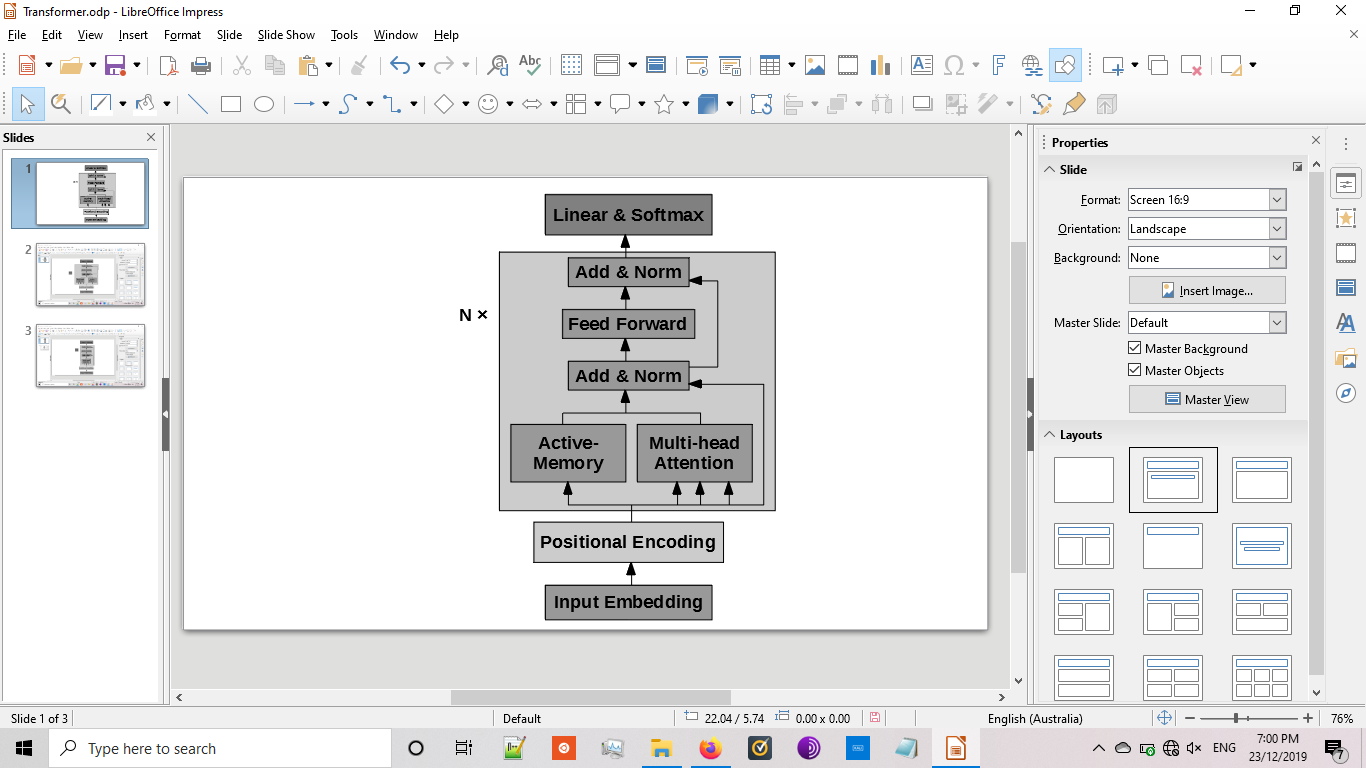}
 \caption{The Self-Attention + Convolutional Operator Transformer.}
\end{center}
\vspace{-6pt}
\end{figure}

\subsection{Self-Attention + Convolutional Operators}

The operator calculates the results of the self-attention mechanism and results of the convolutional operator independently, and then adds them together to produce the final output of the operator. This operator would allow the model to analyze the input using both the self-attention mechanism and active-memory mechanism and decide which features from both mechanisms would be most optimal. This approach has the obvious advantage of being able to take the ‘best of both worlds’, where the optimal features that can only be detected by the self-attention mechanism, and the optimal features that can only be detected by the convolutional operator, are both available to the model.

The architecture of a single layer of the ``self-attention + convolution" operator is shown in Figure 2. This architecture, without the convolutional operator, is a simple Transformer layer. The output of the convolutional operator is added, element-wise, to the output of the self-attention mechanism. This allows, hypothetically, for the best-of-both-worlds, where the model has access to the self-attention mechanism and the active-memory mechanism.

Similarly, we also add the self-attention mechanism to the persistent-convolutional operator and the highway-convolutional operator, respectively.

\section{Experiments}

\begin{table}
\centering
\begin{tabular}{lr}  
\toprule
\textbf{Model}  & \textbf{Loss per Token} \\
\midrule
CGRU       & 1.6834 (+0.1645)    \\
Convolution       & 1.5358 (+0.0169)     \\
Persistent-Convolution    & 1.5341 (+0.0152)    \\
Highway-Convolution   & 1.5327 (+0.0138)    \\
\textbf{Self-Attention}   & \textbf{1.5189 (+0.0)}     \\
Self-Attention + Convolution   & 1.4912 (-0.0277)     \\
Self-Attention + Persistent-Convolution   & 1.4905 (-0.0284)     \\
Self-Attention + Highway-Convolution  & 1.4869 (-0.032)     \\
\bottomrule
\end{tabular}
\caption{The loss-per-token of the self-attention mechanism and the active-memory mechanisms on the WT3 dataset, and the difference of loss between the self-attention and the active-memory mechanisms. The lower the loss, the better the model performed. With the exception of the CGRU, all purely active-memory operators achieve a test loss less then 1.2\% higher then the self-attention mechanism. The optimal models combined the self-attention mechanism and an active-memory mechanism, and achieved a lower test loss than the self-attention mechanism and active-memory mechanisms alone.}
\label{tab:booktabs}
\end{table}

To evaluate the effectiveness of the various convolution-based active-memory mechanisms, we used two separate experiments; a language modelling task that is traditionally associated with attention-based mechanisms \cite{shoeybi2019}, and algorithmic tasks that are associated with active-memory models \cite{kaiser2015}. The active-memory mechanisms are experimented both independently and alongside a self-attention mechanism. 

\subsection{Language Modelling}

\subsubsection{Experimental Setup}
The first task that the operators were tested with was a unidirectional language modelling task; the WikiText-3 (WT3) dataset \cite{merity2016}, tokenized using BPE-tokenization \cite{sennrich2016}. The WikiText-3 dataset was sourced entirely from Wikipedia articles, contains over 3.6 millions lines of text, and is split into a training dataset, valid dataset and test dataset. The train dataset contains 103M tokens, while the valid and test dataset contain 250K tokens each.

The models used were all 8-layer models, with a hidden size of 256 and a filter size of 1024, a vocab size of 32,000, kernel size of 20 and a dropout rate of 0.9. No further regularization was used. The optimizer was the Adam Optimizer \cite{kingma2014} and a warmup-learning rate was used, as specified in \cite{vaswani2017}. All models were implemented using Tensorflow, version 1.07, on a V100 GPU card.

Notable preprocessing was used for analyzing the WikiText-3 dataset; every character was explicitly denoted as lower-case, each hyphen \textit{-} was replaced by \textit{ @-@ } and punctuation marks, such as fullstops and commas, were seperated by white-space. This was done to discourage the BPE to tokenize sets of characters that included punctuation marks, forcing the model to tokenize sets of characters that were only letters, therefore tokenizing a greater set of words.

\subsubsection{Experimental Results}
With the exception of the CGRU operator, all active-memory mechanisms, when combined with the self-attention mechanism, outperformed the self-attention mechanism alone, achieving a lower loss-per-token. This would appear to vindicate the proposition of both this paper and [2], suggesting that, indeed, active-memory mechanisms and self-attention are comparable. However, no model that purely used an active-memory mechanism outperformed the self-attention mechanism for language modelling.

We note that, if the dropout rate was decreased to 0.7, all operators, with the exception of CGRU, all models achieved superior results to the self-attention mechanism at a the same dropout rate. However, these models did not achieve superior results to the self-attention model with a dropout-rate of 0.9. This would imply that self-attention mechanisms are more sensitive to dropout rates compared to active-memory mechanisms.

Further, each operator, except for the CGRU operator, benefited from combining it with self-attention, allowing both operators to operate independently and concurrently. The model with the lowest loss-per-token had a self-attention mechanism and a highway-convolutional operator. It is further worth noting that the highway-convolutional operator outperformed both other convolutional operators, both with and without the addition of the self-attention mechanism.

\subsection{Algorithmic Tasks}

\subsubsection{Experimental Setup}
The second experiment for evaluating the active-memory mechanisms was on various algorithmic tasks:

\begin{table}[!t]
\begin{center}
    \begin{tabular}{c c c c c c c c c c}
    \hline
    (x, y) & 1 & 0 & 1 & 1 & + & 0 & 0 & 1 & 1 \\
    \hline \hline
    (x + y) & 0 & 0 & 0 & 0 & 0 & 1 & 1 & 1 & 0 \\
    \hline
    \end{tabular}
\end{center}
\caption{The binary addition task. Given two numbers (in this case, the two numbers are 11 and 3), the final output is the binary version of the addition of the two input numbers (in this case, 14). }
\end{table}

\begin{itemize}
    \item Reverse: Given an array \textit{X} of size L, the model is trained to return the array \textit{Y}, where \textit{Y[0] = X[-1]}. In order to effectively perform this task, the model must be capable of analyzing the start of the input vector at the very end, and vice-versa.
    \item Sort: Given an array of randomly order integers, the model is trained to return an array that accurately order the input integers. The entire vector must be remembered and analyzed at each time-step.
    \item Addition: Given two binary numbers, the model is trained to return an array that represents the addition in the form of a third binary number. An example of the addition task is shown in Table \textbf{2}.
    \item Multiply: Multiplies two binary numbers, as shown in Table \textbf{3}.
    \item Not: If the input is 1, then not returns 0. Else, the not function returns 1. The output relies only on the input at the current time-step.
    \item Remember: Given a series of random numbers of sequence size $N$, followed by a sequence of zeros of identical size, the model is trained to output a series of zeros of size $N$, followed by the random numbers. In order for this task to be performed, the model must be able to remember tokens over an increasingly long sequence.
\end{itemize}

\begin{table}[!t]
\begin{center}
    \begin{tabular}{c c c c c c c c c c c c}
    \hline
    (x, y) & 1 & 0 & 1 & 0 & 1 & $\times$ & 0 & 1 & 1 & 0 & 0 \\
    \hline \hline
    (x $\times$ y) & 0 & 0 & 0 & 1 & 1 & 1 & 1 & 1 & 1 & 0 & 0 \\
    \hline
    \end{tabular}
\end{center}
\caption{The binary multiplication task. The above example contains two input numbers, 12 and 21, and the output number 252.} 
\end{table}

All data for the algorithmic tasks were generated in an online manner. For three of the tasks, Sort, Addition and Multiply, the model must focus on multiple tokens at every time-step. In comparison, the Reverse task, the Not task and the Remember task only require the model to focus on a single token at every time-step. 

\begin{table*}
\centering
\begin{tabular}{c|c|c|c|c|c|c}
\toprule
\textbf{Model} & \textbf{Reverse} & \textbf{Sort} & \textbf{Addition} & \textbf{Multiply} & \textbf{Not} & \textbf{Remember} \\
\midrule
CGRU & 7.7 & 5.7 & 16.3 & 9.0 & \textbf{104.0} & 9.0 \\
Self-Attention & 41.0 & 14.0 & 7.0 & 7.0 & \textbf{104.0} & \textbf{57.0} \\
Convolution & 17.7 & 20.7 & \textbf{41.0} & \textbf{17.0} & \textbf{104.0} & 36.0 \\
Persistent-Convolution & 25.0 & 20.3 & 38.3 & 16.3 & \textbf{104.0} & 35.0 \\
Highway-Convolution & 19.7 & 16.7 & 35.7 & 13.7 & \textbf{104.0} & 33.0 \\
Self-Attention + Convolution & 41.0 & \textbf{23.3} & 36.3 & 12.3 & \textbf{104.0} & 26.0 \\
Self-Attention + Persistent-Convolution & \textbf{43.7} & \textbf{23.3} & 35.0 & 12.3 & \textbf{104.0} & 27.0 \\
Self-Attention + Highway-Convolution & 41.0 & 20.0 & 34.3 & 11.7 & \textbf{104.0} & 24.7 \\
\bottomrule
\end{tabular}
\caption{The average sequence length that each operator was capable of gaining 100\% accuracy within 100 epochs over 3 runs. The higher the sequence size, the better the model learned. For the Reverse and Sort tasks, the combination of self-attention mechanism and persistent-convolution achieved the best results. For the Sort, Addition, and Multiply tasks, the self-attention mechanism was beaten by the active-memory mechanisms. For the Addition and Multiply tasks, the mere use of a self-attention mechanism alongside an active-memory mechanism actively decreased results. The highest possible sequence that can be learned over these epochs is 104 in the Not task. The self-attention mechanism achieved the best result only for the Remember task. 
}
\label{tab:booktabs}
\end{table*}

The model that was used for algorithmic tasks contains 4 layers, with a hidden size of 128 units, a filter size of 512 and a kernel size of 20. Each model was trained for a maximum of 100 epochs, where each epoch contains 100 iterations. At the end of each epoch, the model was exposed to an online batch, containing 32 test cases. If the model achieved an accuracy of 100\% on the online test batch, the sequence-size of the data is increased, therefore increasing its complexity. 

For the Reverse, Sort, Not and Remember task, when the model achieved a 100\% accuracy, the sequence was increased by 1. For the Addition and Multiply task, the sequence was increased by 2.

The model was initially trained only for sequences that are 5 tokens long and was not introduced to a larger sequence until the model was capable of achieving 100\% accuracy on this sequence-size. We found that this form of curriculum learning was essential: if a model was initially trained on a sequence of several dozen tokens, each operator was incapable of achieving a reasonable accuracy.

The vocabulary size was different for each task. The Reverse task had a vocabulary size of 100, while the Sort task and the Remember task had a vocabulary size of 20. We noted that whenever the vocab size was increased the model would achieve less accurate results. Because all tokens in the Addition, Multiply and Not tasks are either 0, 1, or the separator, the vocabulary size is set to 3.

\subsubsection{Experimental Results}

Each model was tested for each task, and the highest sequence that the model could achieve within 100 epochs was recorded. Each experiment was performed three times, and the average sequence size is presented in Table 4. For example, the self-attention mechanism managed to achieve a 100\% accuracy for a sequence of 41 tokens for the Reverse task, but could not achieve a 100\% accuracy for both the Sort task and the Addition task for a sequence size of 20 (the Sort task achieved a maximum sequence size of 14, while the Addition task achieved a maximum size of 7). 
Of the six algorithmic tasks tested, active-memory mechanisms were used, either solely or in combination with the self-attention mechanism, in the best-performing model of five of these tasks. 
For example, the self-attention mechanism achieved an average sequence size of 41.0 for the Reverse Task and 14.0 for the Sort Task, which are lower than those achieved by the ``self-attention + persistent-convolution" mechanism (43.7 and 23.3, respectively). Furthermore, for the Addition and Multiply Tasks, the active-memory mechanisms across the board outperformed both the self-attention mechanism and the combination of the self-attention mechanism and the active-memory mechanism. For example, the traditional convolution operator, for the Addition Task, outperformed the self-attention mechanism and the ``self-attention + convolutional" mechanism by 34.0 and 4.7 respectively.
The results show that the active-memory mechanisms achieve equal, or superior, results to a traditional self-attention mechanism.

Self-attention, used alone, only performed optimally on the Remember task, and equally well on the Not task.
Interestingly, across all models for the Addition and Multiply tasks, the self-attention mechanism reliably led to poor results; not only does the self-attention mechanism, alone, achieve the poorest results, but the combination of the self-attention mechanism and any active-memory system performed worse then the active-memory system alone. This is in direct contrast to the Sort task and the Reverse task, where the combination of self-attention mechanism and the active-memory achieve the best results.

The self-attention mechanism would, in theory, outperform active-memory mechanisms for the Remember task. This is because, in order to adequately perform the Remember task, the model must be capable of calculating an output based on long-range dependencies, which active-memory cannot match at a large enough sequence length. Other tasks do require a long-range dependency in order to operate well at large sequence sizes, but are dependent on the model performing other tasks as well. For example, the addition task requires to model long range-dependencies and perform binary addition. The self-attention mechanism, although it can learn these long-range dependencies, cannot access all necessary tokens at a given time to adequately perform binary addition. 
This is vindicated by the experimental results. In Table 4, the self-attention mechanism achieved the highest results on the Remember task. This would suggest that, if the algorithmic task only requires a long-range dependency, then the self-attention mechanism will outperform active-memory mechanisms when used alone. In comparison, the self-attention mechanism is incapable of matching the results of active-memory for all other tasks. These findings appear to vindicate the statement made by Kaiser et. al [2]; whenever the sequential task requires the model to focus on multiple tokens at every time-step, using an attention mechanism will lead to extremely poor results, especially in comparison to active-memory models.

It is worth noting that, for each of the active-memory mechanisms operating alone, none of the three achieved a 100\% accuracy for any sequence over a size of 37 for the Remember task. This is because, given the kernel size of 20 and 4 layers, the model is only capable of seeing 37 time-steps across. Therefore, the model cannot see 37 time-steps across and, therefore, cannot perform the Remember task at this sequence size or any larger sequence size. This displays the importance of utilizing both a self-attention mechanism, which can be utilized for analyzing long-range dependencies, and an active-memory mechanism, which can extract features that the self-attention mechanism cannot.

\subsection{Discussion of Results}

The experiments above suggest that, across most tasks, a combination of a self-attention mechanism and an active-memory mechanism, at worst, perform comparably to a purely attention-based model, and at best surpass an attention model, with the exception of the Remember task. However, for some algorithmic tasks, we note that the mere inclusion of a self-attention mechanism actively hinders performance.

Models that combine both the attention mechanism and active-memory mechanisms outperformed both attention-only and active-memory-only models for language modelling. This suggests that, for language modelling tasks, both active-memory mechanisms and attention mechanisms are capable of extracting features that the other mechanism is not capable of extracting, and that both mechanisms operate optimally when used alongside each other.

The findings are further abstracted by studying the effect of various algorithmic tasks; in cases where only a single token needs to be focused on, the self-attention mechanism matches the most ardent active-memory, while active-memory mechanisms radically outperform self-attention for other tasks. This would imply that various time-dependencies that cannot be analyzed by a self-attention mechanism can be analyzed by active-memory.

It is worth noting that, for the Not function, all models learn optimally. This is likely due to the fact that the output of each time-step depends only on the input at this time-step, and each model can analyze this dependency equally efficiently. Also, based on the results of the Remember task, the self-attention mechanism can attain greater long-range dependency in comparison to the active-memory mechanisms.

Finally, we note that, for the Remember function, both mechanisms, when used alone, outperform the two mechanisms used together. For every other task, a combination of the self-attention and active-memory would improve upon at least one of the mechanisms when used alone. We are unsure exactly what has led to this result. This will require further investigation in the future. 

\section{Conclusion}
In this paper we investigate the Transformer’s self-attention mechanism in comparison to a variety of active-memory mechanisms. We experiment on two types of tasks: the language modeling task and the algorithmic task. Our results show that the self-attention mechanism can be improved by an active-memory mechanism alone or by a combination of the two. Our results have implications for wider sequence modeling tasks, which are currently dominated by self-attention based models. 

Our code and models used in experiments are available at: \textbf{\url{https://github.com/Anon-111/Active-Memory}}. 

In the future, we will further explore the use of active-memory for sequence-to-sequence tasks, such as machine translation.
We will also analyze the empirical differences between the studied algorithmic tasks, and investigate why the self-attention mechanism may assist one task but harm another.

\balance
\bibliographystyle{named}
\bibliography{references.bib}

\end{document}